\def\year{2019}\relax
\documentclass[letterpaper]{article} 
\usepackage{aaai19}  
\usepackage{times}  
\usepackage{helvet}  
\usepackage{courier}  
\usepackage{url}  
\usepackage{graphicx}  
\usepackage{algorithm}
\usepackage{algpseudocode}
\usepackage{amsmath}
\usepackage{booktabs,dcolumn,caption}

\begin{document}
%
\title{Plan-Recognition-Driven Attention Modeling for Visual Recognition}
\author{{Yantian Zha, Yikang Li, Tianshu Yu, Subbarao Kambhampati, Baoxin Li}\\
Arizona State University\\
\{yantian.zha, yikangli, tianshuy, rao, baoxin.li\}@asu.edu
}
\maketitle
\begin{abstract}
Human visual recognition of activities or external agents involves an interplay between high-level plan recognition and low-level perception. Given that, a natural question to ask is: can low-level perception be improved by high-level plan recognition? We formulate the problem of leveraging recognized plans to generate better top-down attention maps \cite{gazzaniga2009,baluch2011} to improve the perception performance. We call these top-down attention maps specifically as plan-recognition-driven attention (PRDA) maps. To address this problem, we introduce the Pixel Dynamics Network. Pixel Dynamics Network serves as an observation model, which predicts next states of object points at each pixel location given observation of pixels and pixel-level action feature. This is like internally learning a pixel-level dynamics model. Pixel Dynamics Network is a kind of Convolutional Neural Network (ConvNet), with specially-designed architecture. Therefore, Pixel Dynamics Network could take the advantage of parallel computation of ConvNets, while learning the pixel-level dynamics model. We further prove the equivalence between Pixel Dynamics Network as an observation model, and the belief update in partially observable Markov decision process (POMDP) framework. We evaluate our Pixel Dynamics Network in event recognition tasks. We build an event recognition system, ER-PDN, which takes Pixel Dynamics Network as a subroutine, to recognize events based on observations augmented by plan-recognition-driven attention.   
\end{abstract}

\section{Introduction}
For agents that need to accomplish tasks in challenging dynamic and multi-agent environments, it is important to be able to selectively focus on certain interesting visual regions. For example, an intelligent surveillance camera may monitor different people doing different activities in different areas. This task requires high visual loads. Selective attention \cite{sternberg2016cognitive} allows agents to avoid being overloaded, and make the best use of their intra-agent resources. Thereby, the performance and efficiency is expected to be improved. The attention driven by high-level and task-relevant information is called top-down attention (TDA). TDA in humans generally involves more cognitive processes, consuming mental resources like memory, reasoning, and planning. Selective attention can be driven by top-down feedback, or simultaneously triggered by physical characteristics, using bottom-up attention (BUA). In the field of computer vision, people have traditionally delved into constructing combined BUA and TDA for images, using hand-crafted features \cite{navalpakkam2006}. In recent years, a few works using neural networks to learn combined BUA and TDA models have also emerged, like \cite{anderson2018,girdhar2017,kosiorek17}. Such models may work for static images or video streams. However, none of these has considered if the TDA could be driven by the recognized plans of other agents in the environment, which should be a natural consideration in multi-agent applications. We name this kind of TDA as plan-recognition-driven attention (PRDA). In our work, we address the problem of generating PRDA maps from video frames and corresponding recognized plans.

PRDA may provide practically useful models of selective attention for agents operating in multi-agent environments. In such scenarios, interesting regions far away from an agent are likely to be connected by a plan. And the recognized plan, which is a sequence of actions that an agent is expected to take in the near future, contain useful long-term information for guiding the TDA generation. In this work, we focus on using recognized plans of multiple agents to help generate a PRDA map. We choose event recognition in video surveillance as the primary domain to illustrate the design and evaluation of the proposed method, since surveillance videos often include multiple agents involving in various group activities occurring in various regions of the field of view.

One of the major challenges for using recognized plans is to collect clean observations. This is traditionally done by human or object recognition and tracking. However, in highly cluttered and dynamic environments, methods like object recognition and tracking would become unreliable. For example, humans and objects in the video could be too small to be reliably recognized. Human and object motion can exacerbate the difficulty. Consequently, the collected observations can be incomplete and noisy. To address this problem, we propose a pixel dynamics network (PDN), which learns to directly predict the next state of an object at the position of a pixel. And the prediction is conditioned on both pixel-observation and pixel-level action. Therefore, it works like a dynamics model in control theory \cite{branicky98} -- a model that is able to predict the next state of an object, if given the current state, and an action applied to that object. 

Note that the pixel-level action does not directly belong to a plan, because there is no \textit{intention} for object points. That said, our PDN could decompose a meta action from a plan, to a group of pixel-level actions, which we would explain in detail later. 

To collect meta action features, we extract the Histogram of Optical Flow (HOF), Histogram of Gradients (HOG), and Histogram of Frame Difference (HOD) features from local video frames (video tubes) where an event happens. These video tubes could be obtained by temporally prolonging a bounding box. Then we do clustering on those features, to learn a set of augmented motion primitives (AMPs). Each AMP serves as a meta action, and describes an activity that happens in a video tube.

Our PDN has a subroutine of converting an AMP to pixel-level actions (we denote them as action conditional filters, or ACFs). Then PDN could take a patch of pixels as an observation, and a pixel-level action, to predict the next state of the object point at the center location of that patch. This procedure also has connections to the belief update of the observation model that can handle uncertain observations (i.e., the one in POMDP framework), that we will elaborate later.

To evaluate the effectiveness of PDN, we implement it in an event recognition system. We compare the performance of the event recognition system with our PDN against a state-of-the-art visual recognizer with combined BUA and TDA \cite{girdhar2017}. 


\section{Related Work}
\subsection{Soft Attention Driven by High-Level Signals} 
Here, we briefly introduce previous work on using neural networks to generate soft attention maps, driven by high-level (or top-down) signals. The work \cite{girdhar2017} proposed an attention pooling layer to replace the normal pooling layer in ConvNets. Their high-level signal comes from class labels. For some other previous works that combine top-down and bottom-up attention, their top-down attention model only imitates how spatial reasoning influences attention. In \cite{navalpakkam2006}, they use target object locations and surrounding distractor feature in an image as high-level signals to guide the top-down attention for visual search tasks. In \cite{kosiorek17}, they use the appearance feature of targets. In contrast, our planning-driven attention model exploits both spatial and temporal reasoning for long-term future, based on the awareness of environment dynamics and recognized plan feature of other agents.

\subsection{Pixel-wise Classification}
The design of PDN is also inspired by pixel-wise classification problems (i.e., semantics segmentation). Semantics segmentation is the vision problem that requires assigning a proper semantic label for each pixel. This could be accomplished by using ConvNets \cite{long2015fully,noh2015learning,carreira2012,chen2014semantic,Mostajabi_2015_CVPR}. Our PDN, instead, assigns a “state offset” value for each pixel. 

\subsection{Plan Recognition}
One of the main novelties of our work is to use high-level signals (recognized plans) to form better attention maps. Plan recognition is about predicting the future action sequences of an agent, given its observed actions. While traditionally people rely on classic planning approaches with a domain model to do plan recognition \cite{geffner-ramirez}, recent years have seen an increasing interest in doing plan recognition on a learned and approximated model. Zhuo et al., \cite{dup} treat all plans in a plan library as sentences. Then they use the plan library as the training corpora to train a Word2Vec model. The plan recognition task then becomes to find the action (in the action vocabulary) that is most similar to surrounding actions in a testing plan. The similarity value is provided by the learned Word2Vec. Most recently, \cite{zha2018} extends the problem range of aforementioned work to using action distribution sequences as plans to learn a Distr2Vec model to handle observational uncertainties. 


\section{Our Model}
In this section we introduce our pixel dynamics network, and elaborate how it is used in event recognition. 

\subsection{Augmented Motion Primitives Generation} \label{AMPGen}
Now we introduce our way of generating AMPs as the representation of pixel-level actions. First, we extract features of HOG, HOF and HOD for each frame. Secondly, we do a K-Means clustering on the total features of the whole dataset. We set the number of clusters to $A$. 
We then assign the $\tilde{A}$ nearest centroids of every frame feature. Consequently, we convert a video sequence to an index distribution, which is based on the frame feature clustering. Each distribution has $\tilde{A}$ AMP indices. We also assign a probability value for each index to approximate the uncertainty of an activity. This probability is computed by using the distance between centroids and the corresponding feature. Lastly, we merge several consecutive and identical index distributions to one distribution. This merging of index distribution can make plan traces less redundant. 

\begin{figure}[thpb]
\centering
\includegraphics[scale=0.25]{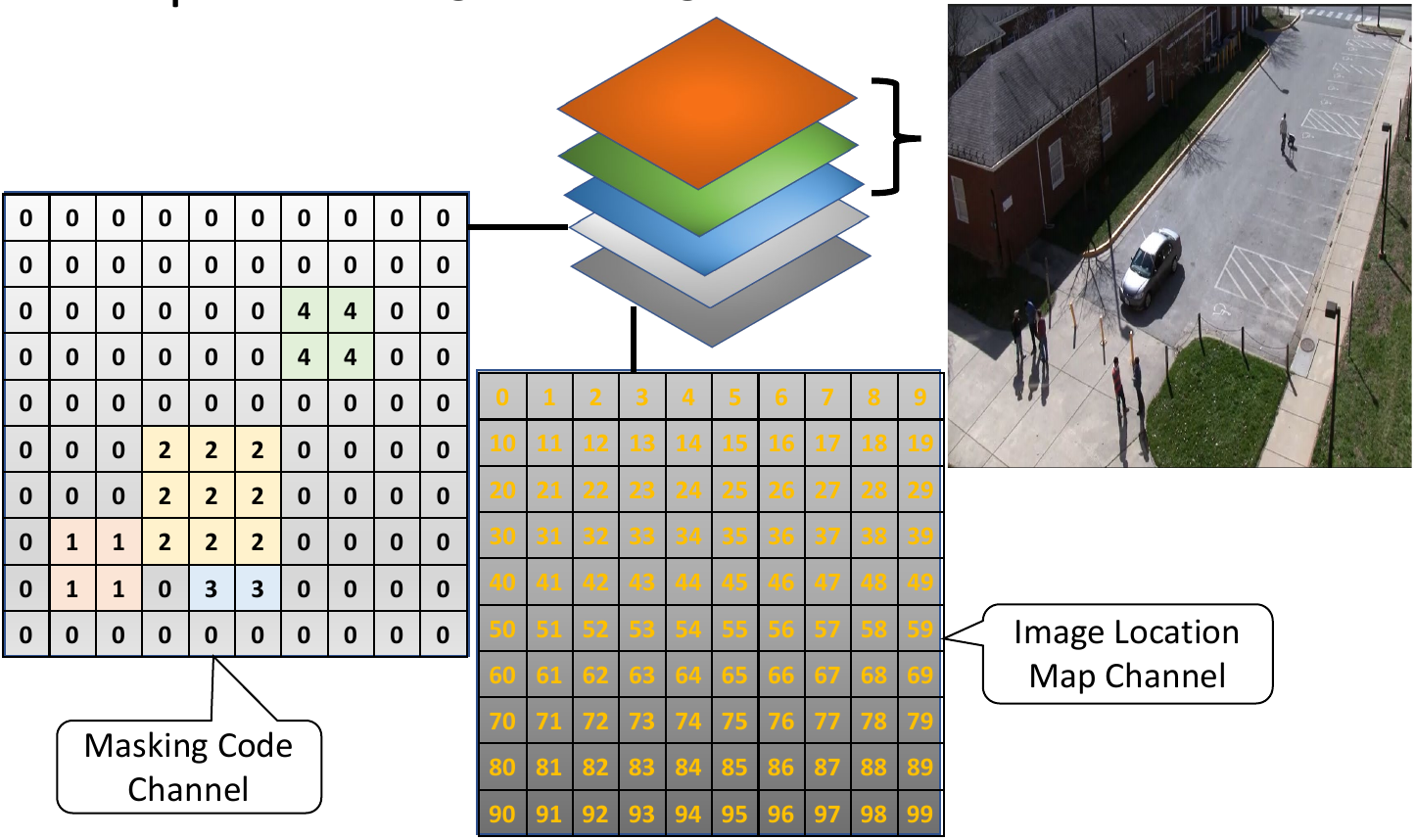}
\caption{Illustration of $V^\mu$, at one step as a PDN input.}
\label{vis_in}
\end{figure}

\begin{figure}[thpb]
\centering
\includegraphics[scale=0.3]{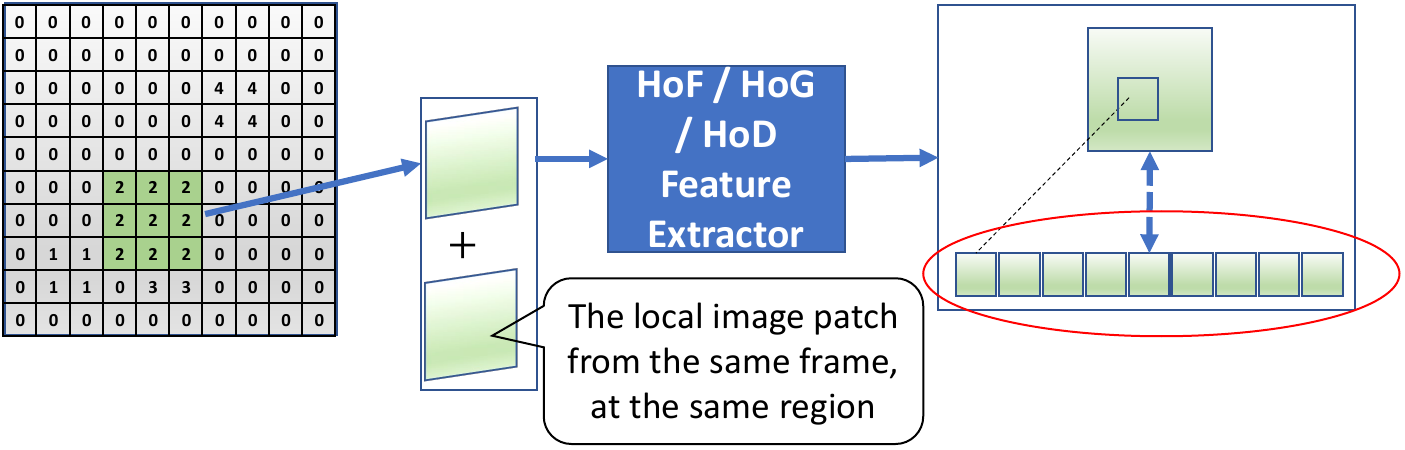}
\caption{Illustration of extracting one AMP (augmented motion primitive) from a local image patch as a PDN input.}
\label{act_in}
\end{figure}

Thus, recognizing the plan of a group of agents would be realized by predicting a sequence of AMPs for a specific region. The region could either by generated by using bounding boxes, or using attention logits as in our work. Then we assign each of these regions, an identical number $m$. This indicates that the $m$-th group agents are doing $m$-th events. If there are totally $M$ such regions, we label each image pixel with those $M$ numbers. This gives an extra channel (channel $c=1$) of masking codes, which serves as a mapping from a recognized plan to the corresponding image area. We also add the second extra channel (channel $c=0$) of pixel locations. After concatenating the two extra channels with original R, G, and B channels, we obtain a masked image, denoted as $V^\mu$, as illustrated in Figure \ref{vis_in}. Correspondingly, the Figure \ref{act_in} illustrates the procedure of using a pair of local images (located by masking channel) to extract one AMP as in the red circle.

\subsubsection{Plan Recognition}
We adopted the work of Distr2Vec, \cite{zha2018}, to generate future AMPs, given a sequence of observed AMP distributions. To elaborate, the plan recognition algorithm is based on a shallow planning model, Distr2Vec (distribution to vector), proposed in the work (Zha et al., 2018). Distr2Vec assumes observed action distribution sequences as inputs. This  is a good fit for using noisy recognition outputs from a vision model , which consists of sequences of distributions over motion primitives. Distr2Vec then learns an action affinity model from the distribution sequences, by minimizing the distance between each of two action distributions. With this trained action affinity model, plan recognition (predicting future actions) is done, by searching for the most similar action (or motion primitive) to observed action distributions.

\subsection{Pixel Dynamics Network}

\begin{figure}[thpb]
\centering
\includegraphics[scale=0.25]{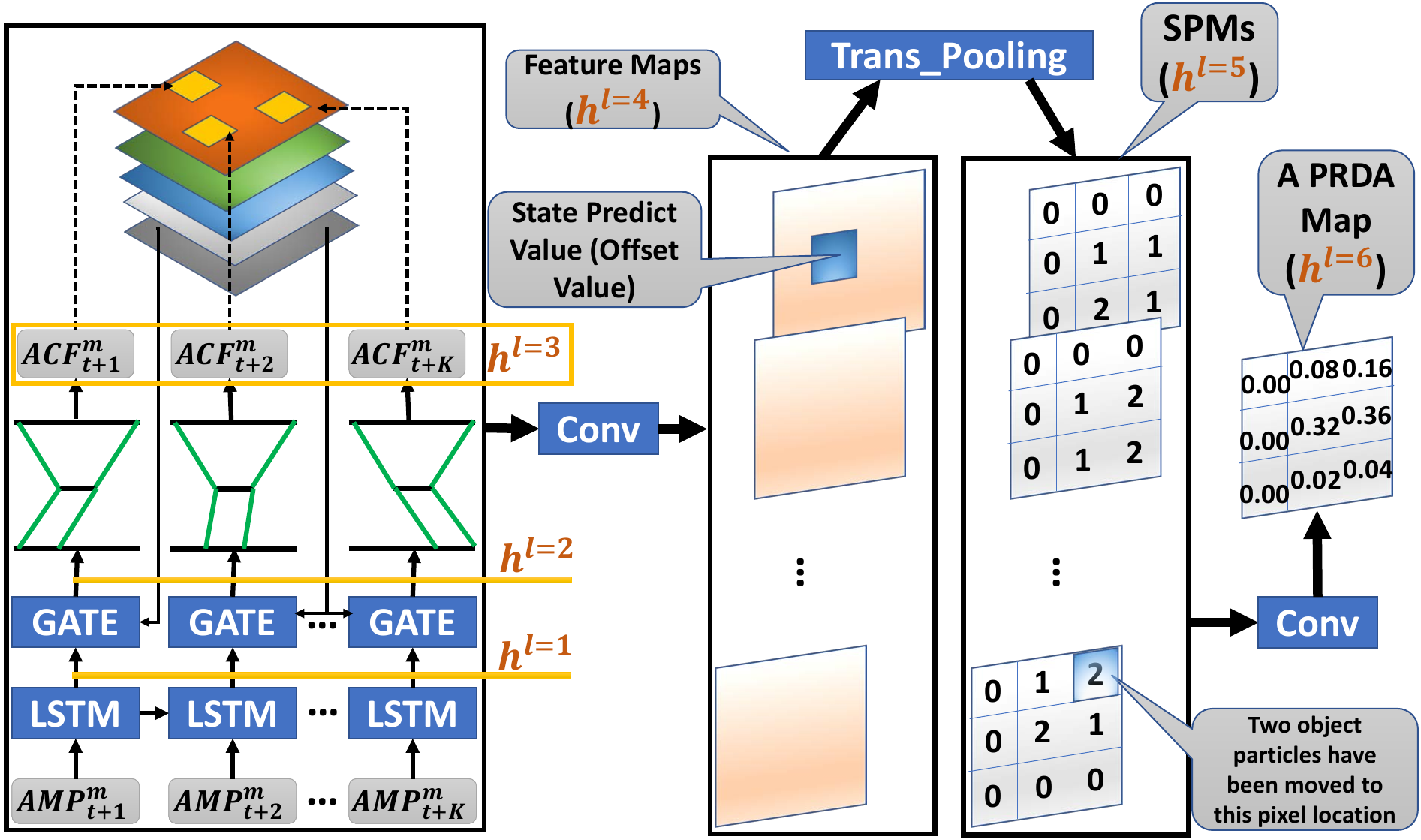}
\caption{The architecture of Pixel Dynamics Network.}
\label{pdn}
\end{figure}

Our PDN also takes both a masked image, $V^\mu$ (Figure \ref{vis_in}), and an AMP (Figure \ref{act_in}), as inputs. With these two inputs, the PDN would generate a state prediction map (SPM). The value at each location of the SPM represents the number of pixels that have been moved to that location by that AMP (action). This is like parallelly applying each of a group of dynamics models to each observation (a local image patch of pixels with corresponding locations) extracted from $V^\mu$. The architecture of PDN is shown in Figure. \ref{pdn}. How PDN works could be divided into three stages: action conditional filter (ACF) generation (hidden output $h^{l=3}$ after third layer), convolution between ACFs and the input image to obtain state prediction values in a set of feature maps ($h^{l=4}$ after fourth layer), and a translation pooling applied to these feature maps, to get a SPM ($h^{l=5}$ after fifth layer). We will introduce the three stages one by one in the following subsections.

\subsubsection{Generation of Action Conditional Filter}
Because we apply our PDN to generate attention maps driven by recognized plans, we feed the input action to a Long Short-term Memory (LSTM) layer. The LSTM layer could handle the sequential correlations among actions in a recognized plan. Note that an AMP contains pixel-level motion information in a local region (of one event). Hence, we cannot directly use the output of (the first) LSTM layer ($h^{l=1}$) as the filter to convolve with a local image patch. The local image patch being convolved might be smaller than a bounding box or a region with high attention logits. As a result, that local image patch only match a sub-region of the input AMP. This is why we have a gate layer. The gate layer is used to select a certain part of $h^{l=1}$ that matches a local image patch, which leads to the need of inputting the channel $c=0$, the pixel location map, to the gate layer. The output of (the second) gate layer, $h^{l=2}$, could be seen as a sparse version of the $h^{l=1}$. After applying the gate function, some parts of $h^{l=2}$ are not usable anymore. When doing convolution between $h^{l=2}$ and a local image patch, some parts in a local image patch would not contribute to the computation. The important information in a local image patch may be missed, which could be a problem. We address this problem by proposing a biased encoding-decoding layer that perform encoding-decoding on $h^{l=2}$, which reflects parts of $h^{l=1}$ selected by the gate layer. Thus, it does encoding (compressing) based on input part that is being paid attention to, and reconstruct with richer details (like zooming in). The reconstruction output per step is an ACF. Generating an ACF is to convert a meta action (AMP) to a group of pixel-level actions. $h^{l=3}$ consists of $M\times K$ ACFs. The equation for using the above procedure to compute an ACF, is shown in Equation. \ref{acf_gen}.

\begin{eqnarray} \label{acf_gen}
ACF^m_{k,t} = bias-decoding(bias-encoding(\nonumber\\LSTM(AMP_{1:k,t}^m) \odot \sigma (W^{GX} V^{\mu,c=0}_{t}+B^G)), \nonumber\\ 1 \leq k \leq K, 1 \leq m \leq M
\end{eqnarray} where $ACF^m_{k,t}$ is the $k$-th action-conditioned filter generated from the $k$-th AMP in a plan of length $K$. That plan is the $m$-th recognized plan, of totally $M$ plans. This is also the plan of $m$-th group of people that are doing $m$-th event. 
We further assume that all of the $M$ recognized plans equally have length $K$. $LSTM(\cdot)$ denotes that the observed $k$ AMPs are fed into a LSTM cell one by one, and output a LSTM state ($h^{l=1}$) per step. $V^{\mu,c=0}_{t}$ is the first channel of input image, containing a map of pixel locations. The values in $V^{\mu,c=0}_{t}$ would range from zero to the product of height of width of input image. The $W^{GX}$ and $B^G$ denote weights and bias parameters of the gate layer. The $\odot$ between LSTM output and gate layer output denotes the element-wise multiplication. The result of this multiplication is fed into a bias-encoding and bias-decoding module. 

Our bias-encoding and bias-decoding module works by performing a K-max pooling on the output from gate layer, which obtain a condensed vector, as shown in the small interval (above the gate module) between two green lines in Figure. \ref{pdn}. Then take that condensed input from K-max pooling as the hidden feature, and adopt normal auto-encoder framework to reconstruct the hidden vector to a feature, which is an ACF.

\subsubsection{Generation of State Prediction Values}
With $K \times M$ generated ACFs, the next task is to use each of them, to convolve with input image. Values in output feature maps after the convolution between an ACF and the input image, would be the offset values of the object particle at the center location of local image patch being convolved with an ACF. The convolution between an ACF and the input image could be described by the Equation. \ref{acf_conv}.

\begin{eqnarray} \label{acf_conv}
h_{ij}^{l=4} = (V^{\mu}_t \odot W^F * F^m_{k,t})_{ij} = \sum_{a=0}^{F_1 - 1} \sum_{b=0}^{F_2 - 1} [[V^{\mu,c=0}_{i+a,j+b} = m]]\nonumber \\\sum_{c=2}^4 (ACF_{a,b}^{m, k} \cdot (V^{\mu}_{c,i+a,j+b} \cdot W^{l=4}_{i+a,j+b}) + B^{l=4}_{i,j}) 
\end{eqnarray} where $h_{ij}^{l=4}$ is the output from convolving ACFs with $V^{\mu}$, i.e., the computation of fourth layer. $W^{l=4}_{i+a,j+b}$ and $B^{l=4}_{i,j}$ are weights and bias in the fourth layer, that is used for the convolution with value at location $(i,j)$. $F_1$ and $F_2$ are respectively the height and width of each ACF. $ACF_{a,b}^{m, k}$ is the value at location $(a,b)$ in the $k$-th ACF of $m$-th plan. As mentioned earlier, $V^\mu_t$ is an masked image at time $t$, with five channels: pixel location map in channel 0, masking code map in channel 1, and RGB maps from channel 2 to 4. Particularly, the masking code map has values that match the index of a recognized plan. In a frame of image, if there are multiple groups of people doing multiple events, then each group matches a highlighted region, and a recognized plan. All pixels in that region are masked by the index of the recognized plan. The values of masking code are in the masking map. If there are $M_t$ different recognized plans at time $t$ (as shown in Figure. \ref{overall_er}, there would be $M_t+1$ different values in masking map. The reason is, we assign the value zero, to the regions of pixels do not belong to any of recognized plans.  

\subsubsection{Translation Pooling}
Once obtaining a feature map of state prediction values, the next step is to convert that to a SPM, which is accomplished by our translation pooling layer. A state prediction value at a position represents the offset as explained before, and the translation pooling layer works by converting that offset value, to a "plus one" operation at a new position, after applying the offset value to the old position. The algorithm of translation pooling is shown in Algorithm. \ref{alg_tp}.



\begin{algorithm}
  \caption{Translation Pooling}\label{alg_tp}
  \begin{algorithmic}[1]
    \Procedure{trans-pooling}{$h^{l=4}, shape (N, N)$}  \Comment{From a set of $N \times N$ feature maps, in which values are computed using  Equ. \ref{acf_conv}}

  	 \State $h^{l=5}_{0:N\times N-1}\gets 0$\Comment{Initialize all values in the vector $h^{l=5}$ (size is $N^2$) to zeros}

      \For{\texttt{$i=1; i \leq N; i++$}}
      \For{\texttt{$j=1; j \leq N; j++$}}
     	 \State $h^{l=5}_{ID(i,j)-h^{l=4}_{ij}} += 1 $\Comment{Generate statistics using all state prediction values in $h^{l=4}$}
      \EndFor
      \EndFor
      \State $SPM \gets reshape \: h^{l=5}: from\: (N^2)\: to\: (N,N)$
      \State \textbf{return} $SPM$\Comment{Return a state prediction map}
    \EndProcedure
    \Procedure{id}{$i, j$}\Comment{Convert 2D index to 1D index}
      \State \textbf{return} $i*N+j$
    \EndProcedure
  \end{algorithmic}
\end{algorithm}


\subsection{PRDA Generation Layer}
The role of this layer is to pixel-wise summarization across all channels of SPMs (feature maps) from the translation pooling layer. This summarization will be taken as a PRDA map. Intuitively, as each of those SPMs maps to an action of an agent, summarization of SPMs would provide useful information of which visual region is likely to get considerable amount of pixel objects. As mentioned before, a pixel object means the object point at a location with the corresponding pixel as the observation. The PRDA generation layer also needs to output a feature map that has the same size of a SPM and $V^{\mu}$. We design a special ConvNet layer with filter size, weight, and slide stride set to one. And the weights in this layer are not trainable. This way, we are able to do summarization over all pixel regions, and across all SPMs, which is identical to Equation \ref{PRDA_Gen}.

\begin{eqnarray} \label{PRDA_Gen}
h_{ij}^{l=6} = PRDA^{t+1}_{i,j} = \frac{\sum_{c=0}^{M \times K-1} SPM^{t,i,j}_{c}}{Z}
\end{eqnarray} where $Z$ is a normalization number, and we set it to one in our experiments. In $h_{l=5}$, there are $M \times K$ SPMs, because there are $M$ plans of length $K$, and each AMP leads to a SPM. Each SPM has $N \times N$ values. Generating a PRDA map requires calculating statistics over all of them. In addition, here $c$ denotes the $c$-th channel or SPM in $h_{l=5}$, $t$ denotes the current time step, and each pixel location in a PRDA map is identified by $i$ and $j$.

\subsection{Event Recognition with PDN}
\begin{figure}[thpb]
\centering
\includegraphics[scale=0.28]{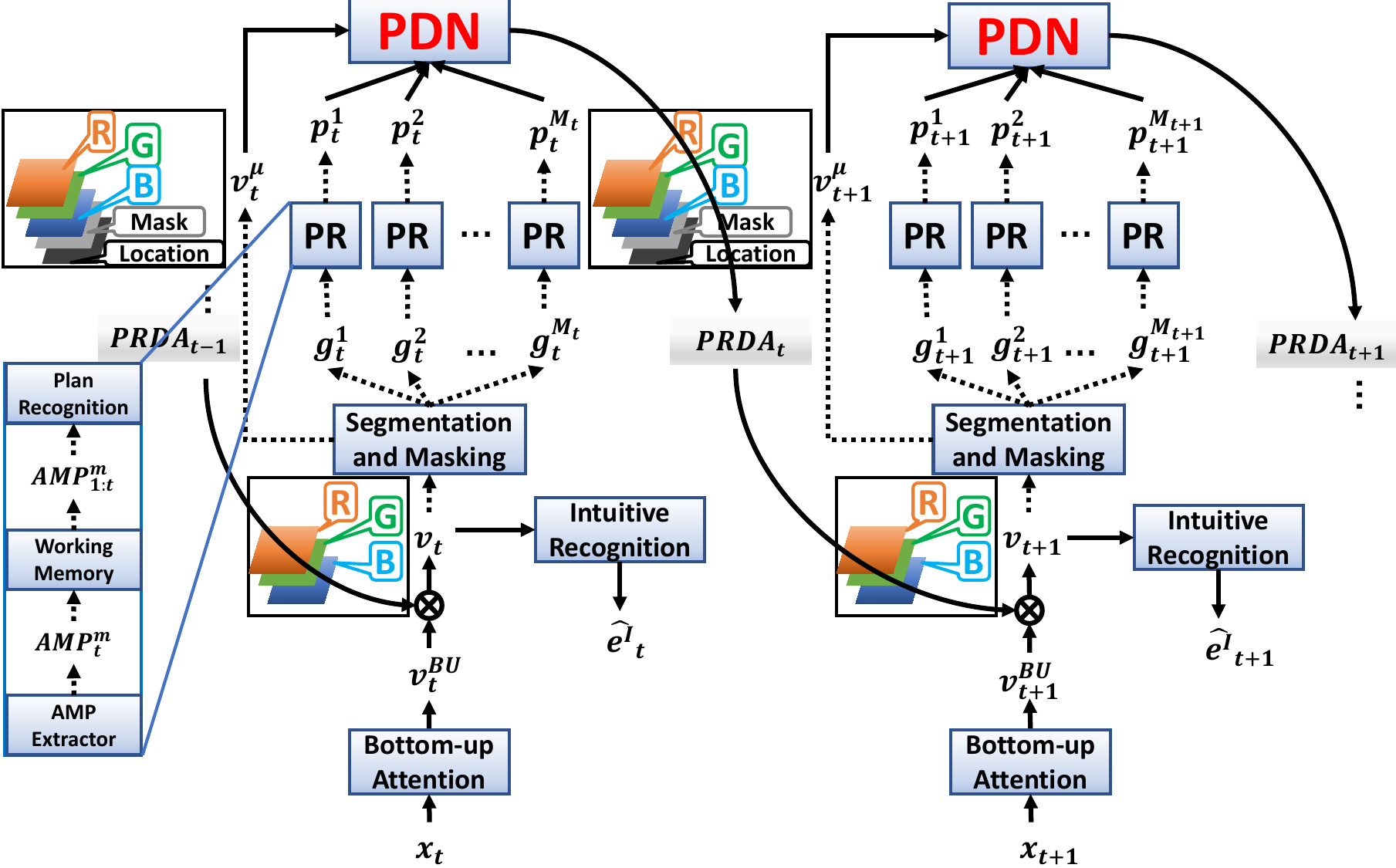}
\caption{The event recognition system with Pixel Dynamics Network.}
\label{overall_er}
\end{figure}

Here we show how our PDN can be used to generate PRDA maps to support an event recognition task. We denote our event recognition system as ER-PDN. The overall framework is shown in Figure. \ref{overall_er}. The solid paths are differentiable, and dashed ones are not. At each step, a video frame $X_t$, is filtered by bottom-up attention model, to get a BUA map $V^{BU}_{t}$. The $V^{BU}_{t}$ is then combined with a PRDA map from the previous step, to get a highlighted frame $V_t$. The segmentation and masking component then extracts $M_t$ glimpses, which are disconnected highlighted areas in $V_t$. Each of these glimpses are fed into a separate but structurally identical plan recognition component, PR, to generate $M_t$ recognized plans $p$. The segmentation and masking component also adds additional channels as explained before, to map each of $M_t$ recognized plans that is indexed by $m$, to a highlighted region in $V_t^\mu$. Consequently, every pixel in $V_t^\mu$ also has a mask code $m$. The PDN takes all recognized plans for different highlighted areas, and the masked $V_t^{\mu}$, to generate the PRDA map $PRDA_{t}$. 

The above procedure describes the non-differentiable path (the dashed ones). In order to train the model, we want to let the model to predict a label so that there is a teaching signal. This is what the dashed path is showing. After generating $V_t^{\mu}$, an intuitive recognition module is required to predict an event label at each step. The intuitive recognition module is a LSTM layer, which has the advantage of handling sequential inputs with internal long-term dependency. 
  
\section{Deeper Analysis of PDN}
\subsection{Connection to Belief Updates in Observation Models}
We start from providing the belief update procedure in the observation model, in partially observable Markov decision process (POMDP) framework \cite{monahan1982state,littman2009tutorial}. Observation model could handle observation uncertainties, and model the evolution of states. POMDP framework assumes state uncertainty, and treats actions as triggers between states (the essential difference to belief updates in other observation models, e.g., HMMs). The belief update process in POMDP is described in Equation \ref{eq_init_bel}, \ref{eq_o}, \ref{eq_T}, and \ref{eq_bel_upd}. In Equation \ref{eq_init_bel}, $b_0(s)$ denotes the initial belief of state $s$. Equation \ref{eq_o} and \ref{eq_T} define the observation model and transition model respectively. Equation \ref{eq_bel_upd} defines how belief is updated.  
\begin{gather}
b_0(s) = Pr(S = s) \label{eq_init_bel}\\
O(s', a, o) = Pr(o|s', a) \label{eq_o}\\
T(s, a, s') = Pr(s'|s, a) \label{eq_T}\\
b'(s') \propto O(s', a, o) \sum_{s\in S} T(s,a,s')b(s) \label{eq_bel_upd}
\end{gather}

Now we elaborate the connection between our PDN and POMDP's observation model. PDN takes an image and an action (later becomes to an ACF) as inputs, and predict how each object point at each image location (each location maps to a pixel value) would be shifted. The state $s$ denotes which object point at which image location. We assume this state is hidden. We can only guess this state from the observation of a group of pixels. Thus the observation uncertainty in PDN essentially means: given the observation of a group of pixels (e.g., a square patch of pixels), the true location of the object point that should be in the center of that patch, could actually be at anywhere in that patch. 

Equation \ref{acf_conv}, which predicts location offset values of a certain object point ($h_{ij}^{l=4}$), works like using a ConvNet to model the belief updates (Equation \ref{eq_bel_upd}). Equation \ref{eq_o} and \ref{eq_T} are subroutines of Equation \ref{eq_bel_upd}. To predict the $s'$ in Equation \ref{eq_T}, the neural network that approximates $T(s,a,s')$ needs the inputs of an image location with the corresponding pixel value, and the corresponding value in an ACF. If we multiply a value from an ACF with a pixel value in the image location $(i,j)$, we are assuming (or believing) that that pixel is the proper representation of the object point at location $(i,j)$. Thus $T(s,a,s')$ in POMDP matches the entry of $\sum_{c=2}^4 (ACF_{a,b}^{m, k} \cdot (V^{\mu}_{c,i+a,j+b} \cdot W^{l=4}_{i+a,j+b}) + B^{l=4}_{i,j}$ in Equation \ref{acf_conv} (state $s$ matches $V^c_{i+a,j+b}$, and action $a$ matches $ACF_{a,b}^{m, k}$). The summarization over states, $\sum_{s\in S}$, in Equation \ref{eq_bel_upd} (belief updates in POMDP), matches $\sum_{a=0}^{F_1 - 1} \sum_{b=0}^{F_2 - 1}$ in Equation \ref{acf_conv}. These three matches intuitively mean that: the PDN believes that all image locations in a local image patch (to be convolved with an ACF), and the corresponding pixel value could represent the true object point state with certain probabilities. Also, note that when PDN predicts the offset value, PDN is not really moving that object point, but is only observing. Hence, there is no uncertainty in whether an object point is really moved to the target image location or not. In other words, $T(s,a,s')$ is deterministic, and $O(s',a,o)=1$ (for Equation \ref{eq_o}). If we further assume that the belief of states obeys to uniform distribution, i.e., all locations in a local image patch are equally possible to be where the main object point is, then $b'(s') \propto \sum_{s\in S} T(s,a,s')$, which can be reduced to solving the Equation \ref{acf_conv}.

\subsection{Relationship with Other Models that can do Pixel Transformation}
\subsubsection{Spatial Transformer Networks (STN)}
In the work of STN \cite{jaderberg15}, a localization network figures out a set of proper transformation parameters, with the input of an image. These parameters are fed into a grad generator and sampler to transform feature maps. Thus, pixel-level transformation leads to image-level transformation, like rotation, crop, etc. Our proposed PDN shares a key characteristics with STN, with respect to learning to manipulate pixels (pixel transformations). That being said, STN’s pixel transformation has a different purpose than PDN, which leads to different designs of model architectures and algorithms. STN intends to warp an entire image, which requires that all pixels in new feature maps still maintain proper values such that they all together form a transformed image. However, for PDN, the purpose is to figure out which image position has been moved a comparable number of object points given pixel-level actions. This statistics indicates proper attention importance. The hidden layer for generating SPMs treats pixels with their corresponding image locations, as observations, and object point with a location, as states. That hidden layer generates a set of state offset values, each reflects where the next image position of an object point is believed to be (belief update). 

\subsubsection{Dynamic Filter Networks} 
The design philosophy of DFNs \cite{jia16} is coincidently similar to the ACF generator in PDN: instead of learning filter weights in ConvNets, they have a neural network model, trained to generate values of filter weights on the fly. Thus, DFNs can be seen as a more general version of ACF generator. And the ACF is specifically used to generate tuned pixel-level action features (AMPs), for each image region.

\section{Evaluation}
In this section, we provide experiment results and analysis. We compare our PDN with a baseline vision model \cite{girdhar2017}, by using them to solve an event recognition task. Note that the  baseline model also has an attention layer. Unlike existing event recognition works which only use event-relevant sequences \cite{wang2015video,mahmud2017joint}, we did not do pre-segmentation or use bounding boxes. We ran our experiments on a machine with a Quad-Core CPU (Intel Xeon 3.4GHz), a 64GB RAM, a GeForce GTX 1080 GPU, and Ubuntu 16.04 OS.

\subsection{Dataset Pre-processing and Training Procedure}
We used the VIRAT 1.0 Ground Dataset \cite{oh2011large} for evaluation. We randomly select 70\% of all videos for training, and the rest for testing. And we down-sampled videos to use every six frames. We used different pre-processing procedures for learning augmented motion primitives (AMPs), and learning the PDN. 

To learn motion primitive clusters, we applied K-means clustering to find the motion primitive cluster library. The motion primitive cluster library is a library of mappings between a motion primitive vector, and a cluster index. We set the number of clusters to 512, and apply clustering on HoD features collected from 70\% of VIRAT videos for training. We did not use the original VIRAT videos. Instead, we used the video segments that are inside ground-truth bounding boxes. This means that the videos we use are both temporally and spatially smaller than the original videos. Those bounding boxes highlight event relevant frames and objects. In contrast, when training the ER-PDN, we did not use any bounding box information in videos, which significantly increases the event recognition difficulty. Once we obtain the AMP clusters, we can convert sequences of HoD features to a sequence of motion primitive (or AMP) indices. Each of such a motion primitive index sequences could be seen as a high-level plan. Then we use the work of \cite{zha2018} to learn a shallow plan recognition model. When training the plan recognition model, we set the distribution size to three, and training epoches to sixty.

When training ER-PDN and the baseline model \cite{girdhar2017}, We feed both models with resized images ($450 \times 450$). Note that if there are multiple events simultaneously happening, then some video frames may map to multiple labels. For this situation, we copy certain images multiple times, and assign each with a corresponding event label. The video down-sampling allows us to increase the variance between each of two frames in our pre-processed dataset, and thereby increase the variance between each of two AMPs in a recognized plan.

When we train the ER-PDN, we feed four frames as a batch to the pre-trained plan recognition module to generate a $K$-step recognized plan (we set $K$ to five). With two consecutive frames, the HOD feature extraction component generates a HoD feature vector, which would be used for searching top-3 AMP indices from the trained AMP cluster library. Thus, we could obtain a sequence of three AMP index distributions (each distribution has three AMPs), given an input image batch. This is similar to the work \cite{zha2018}. This AMP distribution sequence is used as an observed plan, fed into the plan recognition module, which returns a $k$-step sequence of future AMP indices that is most likely to happen (the recognized plan). ER-PDN then converts the future AMP indices back to a sequence of motion primitive feature vectors, by using the AMP cluster libraray. ER-PDN then feeds these vectors one by one, to generate proper ACFs, and SPMs. This is how ER-PDN uses input actions (AMP vectors). 

To generate a PRDA, ER-PDN also takes an image as input. After taking a batch of four consecutive images from a video, ER-PDN uses the first image to generate a BUA, which is then used to generate glimpse for plan recognition. And ER-PDN uses the last image, with the generated PRDA and BUA map, to augment the input and predict which event is happening.


\subsection{Results and Analysis of Event Recognition}

\begin{figure*}[tb] 
\centering
 \makebox[\textwidth]{\includegraphics[width=.55\paperwidth]{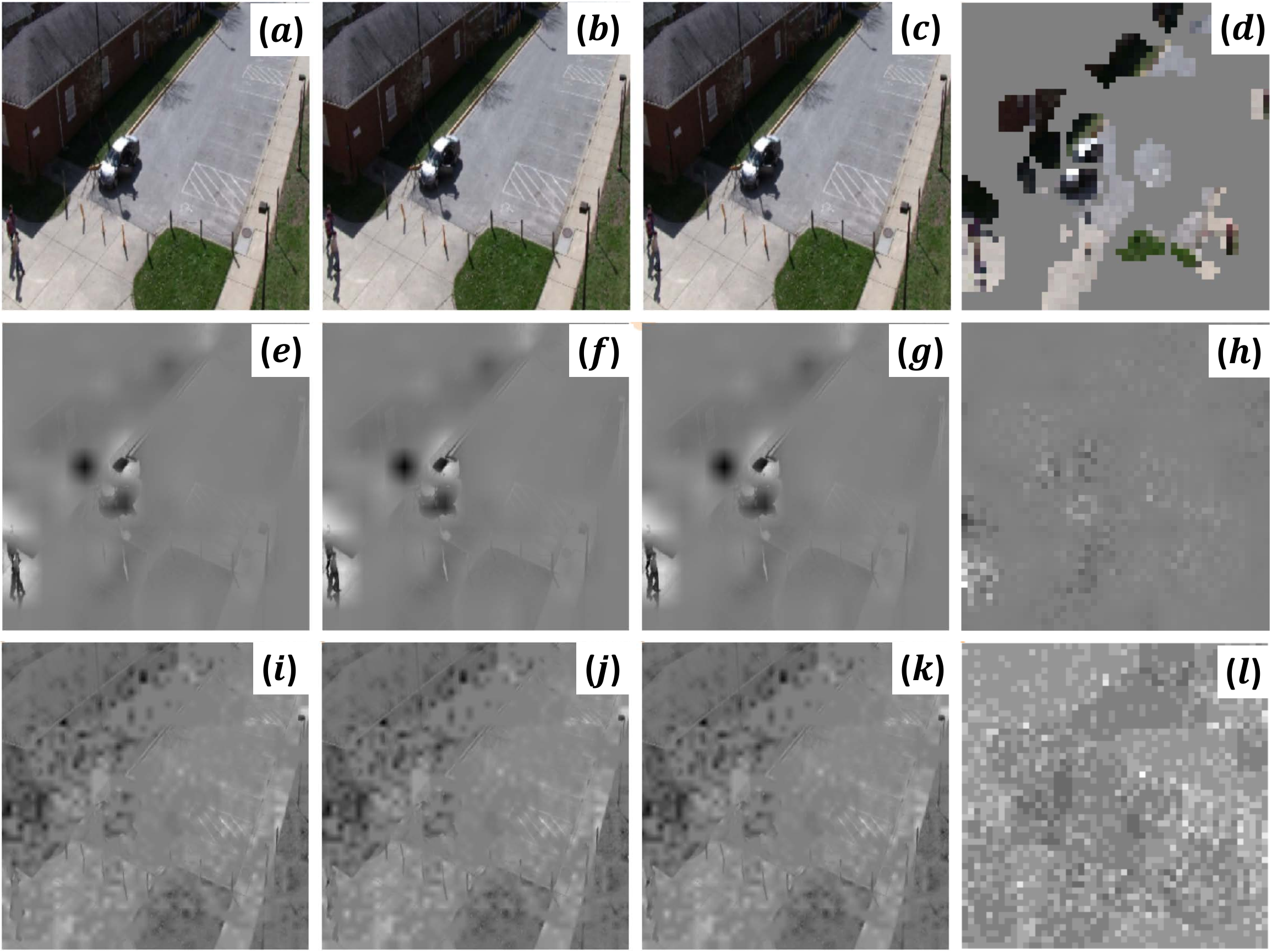}}
 \caption{Results Visualization of our Event Recognition System with the PDN.}

\label{fig:eval_1}
\end{figure*}


\begin{table}[h]
\begin{center} {\footnotesize
\begin{tabular}{cc}
\hline
 Model 
 & Mean Average Precision\\
\hline
Baseline \cite{girdhar2017} & 29.06\% \\[0ex]
ER-PDN & 29.15\%\\[0ex] 
\hline
\end{tabular} }
\end{center}
\caption{\footnotesize Statistics of event recognition performance on VIRAT dataset, using the baseline model \cite{girdhar2017}, and our ER-PDN.}
\label{tbl_result}
\end{table}

A comparison of the results in Table \ref{tbl_result} shows that the visual recognition could be improved by a plan (a sequence of motion primitives). In addition, Figure \ref{fig:eval_1} contains examples of input video frames, and the corresponding feature maps of BUA, PRDA, and $V_t^{\mu}$. Specifically, in Figure \ref{fig:eval_1}, ($a$), ($b$) and ($c$) are resized video frames. ($d$) is a visualization of $V_t^{\mu}$, by applying a BUA to a resized frame and keeping parts within a threshold. ($e$), ($f$), and ($g$) are visualizations of applying a BUA to a resized frame. ($h$) is the feature map of applying both a BUA and a PRDA to the output of a convolutional layer, with an input frame. ($i$), ($j$) and ($k$) are visualizations of applying a PRDA to input video frames. ($l$) is a PRDA feature map. 

We can observe from ($a$), ($b$) and ($c$) in Figure \ref{fig:eval_1} that, three people are moving toward a car. ($e$), ($f$), and ($g$) in Figure \ref{fig:eval_1} indicate that, the BUA map focus properly to interesting image areas, where there is a car, and people. Because the car is not moving, in ($i$), ($j$) and ($k$), the area with car is darker, which indicates less attention. The area between the three people, and the car, is lighter, which suggests that the moving people may want to get to the car. Thus, the lighter region between the people and the car indicates that there is more attention put there. 

\section{Conclusion and Future Work}
In this work, we formulated the problem of generating attention maps which are driven b plans. While we assume that the plans come from a plan recognition module, the plans could also come from, e.g., an agent itself. Essentially, those plans contain long-term information that an agent could make use of, in order to focus on proper regions that are closely relevant to a task.

To address this problem, we proposed a Pixel Dynamics Network (PDN) that learns a dynamics model for object points in video frames. We assume that, in each pixel location, there is an object point, and PDN learns to predict the next pixel location (as the next state) of that object point, on an observation of several pixels, and a motion primitive vector (as an action). Furthermore, if we feed PDN with a sequence of motion primitives (i.e., a plan), then PDN could predict a sequence of future states. If we apply PDN to all object points that a video frame could reflect, and all actions from one or more plans, we could obtain a set of feature maps, that show a statistics of $K$-step future states of all object points. This set of feature maps are called state prediction maps (SPMs). By summarizing them, we could obtain the plan-recognition-driven-attention map (PRDA). We empirically show that an event recognition task could be improved by using PRDA maps, i.e., the plan-recognition-driven-attention modeling. 

In future work, we plan to leverage not only HoD features, but HoF, and HoG feature as well, to learn better motion primitive features, to improve PDN performance. 

\section{Acknowledgements}
This research is supported in part by the AFOSR grant FA9550-18-1-0067, the ONR grants N00014161-2892, N00014-13-1-0176, N00014- 13-1-0519, N00014-15-1-2027, and the NASA grant NNX17AD06G. 

\bibliographystyle{aaai.bst}  
\bibliography{output}  

\end{document}